\documentclass[journal]{IEEEtran}
\usepackage{latexsym}

\usepackage{enumerate}
\usepackage{graphicx}
\usepackage{subfigure}
\usepackage{multirow}

\usepackage{framed,multirow}
\usepackage{times}
\usepackage{graphicx} % more modern
\usepackage{subfigure}
\usepackage{amssymb}
\usepackage{latexsym}

\usepackage{algorithm}
\usepackage{algorithmic}

\usepackage{hyperref}

\usepackage{url}
\usepackage{xcolor}
\usepackage{amsmath}
\usepackage{amsxtra}
\usepackage{braket}
\usepackage{array}
\usepackage{color}
\usepackage{url}
\usepackage{placeins}

\usepackage{times}
\usepackage{balance}
\usepackage{enumerate}
\usepackage{multirow}
\usepackage{pifont}
\usepackage{marvosym}
\usepackage{ifsym}
\usepackage{threeparttable}

\begin{document}
\title{QESK: Quantum-based Entropic Subtree Kernels for Graph Classification}
\author{Lu~Bai,~\IEEEmembership{}
        Lixin~Cui,~\IEEEmembership{}
        and~Edwin R.~Hancock,~\IEEEmembership{IEEE~Fellow}

\thanks{Lu Bai (bailu@bnu.edu.cn) is with School of Artificial Intelligence, Beijing Normal University, Beijing, China, and Central University of Finance and Economics, Beijing, China. Lixin Cui (${}^{*}$Corresponding Author: cuilixin@cufe.edu.cn), and Edwin R. Hancock (edwin.hancock@york.ac.uk) is with the Department of Computer Science, University of York, York, UK}% <-this % stops a space
}
\markboth{IEEE Transactions on ...}%
{Shell \MakeLowercase{\textit{et al.}}: Bare Demo of IEEEtran.cls for Journals}

\maketitle

\begin{abstract}
%\bol\begin{abstract}
  In this paper, we propose a novel graph kernel, namely the Quantum-based Entropic Subtree Kernel (QESK), for Graph Classification. To this end, we commence by computing the Average Mixing Matrix (AMM) of the Continuous-time Quantum Walk (CTQW) evolved on each graph structure. Moreover, we show how this AMM matrix can be employed to compute a series of entropic subtree representations associated with the classical Weisfeiler-Lehman (WL) algorithm. For a pair of graphs, the QESK kernel is defined by computing the exponentiation of the negative Euclidean distance between their entropic subtree representations, theoretically resulting in a positive definite graph kernel. We show that the proposed QESK kernel not only encapsulates complicated intrinsic quantum-based structural characteristics of graph structures through the CTQW, but also theoretically addresses the shortcoming of ignoring the effects of unshared substructures arising in state-of-the-art R-convolution graph kernels. Moreover, unlike the classical R-convolution kernels, the proposed QESK can discriminate the distinctions of isomorphic subtrees in terms of the global graph structures, theoretically explaining the effectiveness. Experiments indicate that the proposed QESK kernel can significantly outperform state-of-the-art graph kernels and graph deep learning methods for graph classification problems.
\end{abstract}


\maketitle
\IEEEpeerreviewmaketitle

\section{Introduction}\label{s1}

In structure-based pattern recognition, graphs are natural and powerful representations that can model pairwise relationships between components of complex systems~\cite{DBLP:journals/ftml/BorgwardtGLOR20}. Typical instances include the graph structures abstracted from (a) digital images~\cite{DBLP:journals/jmiv/BaiH13}, (b) 3D shapes~\cite{DBLP:conf/cvpr/EscolanoHL11}, (c) molecule structures~\cite{DBLP:conf/nips/GasteigerBG21}, (d) transportation systems~\cite{DBLP:journals/apin/ZengPHYH22}, etc. One fundamental challenge arising in graph data analysis is how to learn meaningful numerical features of discrete graph structures for the objective of graph classification. To this end, one popular way is to utilize graph kernels and directly compute the similarities between graphs in a potential high-dimensional Hilbert space. More specifically, this not only helps us to better preserve the structural information of graphs in the Hilbert space, but also provides a straightforward way to employ standard kernel-based machine learning algorithms for graph clustering or classification, e.g., the kernel-based Principle Component Analysis (kPCA)~\cite{DBLP:journals/neco/ScholkopfSM98}, $K$-Nearest Neighbor (KNN)~\cite{DBLP:series/smpai/RiesenB10}, C-Support Vector Machine (C-SVM)~\cite{DBLP:journals/pr/NeuhausB06}, etc. The aim of this work is to propose a novel quantun-inspired graph kernel. The proposed kernel is based on constructing entropic subtree representations associated with the Continuous-time Quantum Walk (CTQW)~\cite{DBLP:journals/pr/BaiH14B}. These entropic subtree representations can provide a powerful way to represent the complicated intrinsic characteristics of graph structures.

\subsection{Literature Review}\label{s1.1}
%\begin{align}
%F_p=\{\mathrm{N}_p(s_1),\ldots,\mathrm{N}_p(s_x),\ldots, \mathrm{N}_p(s_{|\mathbf{S}|}) \},
%\end{align}
%and
%\begin{align}
%F_q=\{\mathrm{N}_q(s_1),\ldots,\mathrm{N}_p(s_x),\ldots,  \mathrm{N}_q(s_{|\mathbf{S}|}) \},
%\end{align}

In machine learning and pattern recognition, a graph kernel is defined as a similarity measure between a pair of graphs. This measure should be symmetric, well defined, and positive definite. In fact, most of the state-of-the-art graph kernels are defined based on the principle of R-convolution, that was first introduced by Haussler in 1999~\cite{haussler99convolution}. This is a generic principle of defining new graph kernels by decomposing the global structures of graphs into local substructures and measuring the isomorphism between the substructures.

\begin{figure*}[t]
\centering
\subfigure{\includegraphics[width=1.0\linewidth]{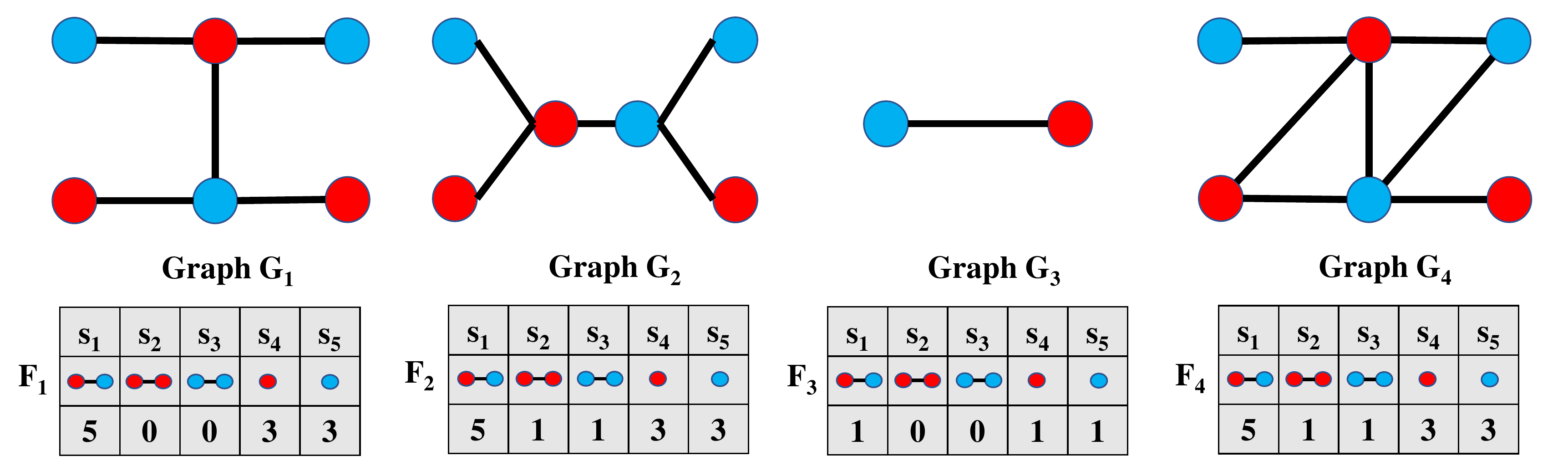}}
\vspace{-15pt}
\caption{Example graphs where different colors correspond to specific vertex attributes} \label{ExampleG}
\vspace{-15pt}
\end{figure*}

Specifically, for a set of graphs $\mathbf{G}$, assume $G_p(V_p,E_p)\in \mathbf{G}$ and $G_q(V_q,E_q)\in \mathbf{G}$ are a pair of sample graphs, and ${S}_p=\{{s}_{p;1},\ldots,{s}_{p;x},\ldots,{s}_{p;|V_p|}\}$ and ${S}_q=\{{s}_{q;1},\ldots,{s}_{q;y},\ldots,{s}_{q;|V_q|}\}$ are their sets of substructures respectively, based on a specific graph decomposing method $\mathrm{F}: G_p \rightarrow S_p$. A standard R-convolution graph kernel $\mathrm{K}_{R}$ between $G_p$ and $G_q$ can be formulated as
\begin{align}
\mathrm{K}_\mathrm{R}(G_p,G_q)=\sum_{x=1}^{|V_p|} \sum_{y=1}^{|V_q|} \delta(s_{p;x},s_{q;y}),\label{RK}
\end{align}
where $s_{p;x}\in S_p$, $s_{q;y}\in S_q$, and $\delta(s_{p;x},s_{q;y})$ is defined as
\begin{equation}
\delta(s_{p;x},s_{q;y})=\left\{
\begin{array}{cl}
1   & \mathrm{if} \  s_{p;x} \simeq s_{q;y}),\ \\
0   & \mathrm{otherwise},
\end{array} \right. \label{DiracK}
\end{equation}
where $\delta$ is usually a Dirac kernel. Here, $\delta(s_{p;x},s_{q;y})$ is equal to $1$ if the substructures $s_{p;x}$ and $s_{q;x}$ are isomorphic to each other (i.e., $s_{p;x} \simeq s_{q;y}$), and $0$ otherwise. Clearly, the kernel value of $\mathrm{K}_\mathrm{R}$ corresponds to the number of isomorphic substructure pairs between $G_p$ and $G_q$. Thus, Eq.(\ref{RK}) can be re-written as a dot product formula, i.e.,
\begin{align}
\mathrm{K}_\mathrm{R}(G_p,G_q)=\langle F(G_p), F(G_q)  \rangle,\label{RK_PD}
\end{align}
where $F(G_p)=\{\mathrm{N}_p(s_1),\ldots,\mathrm{N}_p(s_x),\ldots, \mathrm{N}_p(s_{|\mathbf{S}|}) \}$
and $F_q=\{\mathrm{N}_q(s_1),\ldots,\mathrm{N}_p(s_x),\ldots,  \mathrm{N}_q(s_{|\mathbf{S}|}) \}$
are the feature vectors of $G_p$ and $G_q$, each of their elements $\mathrm{N}_p(s_x)\in F_p$ and $\mathrm{N}_q(s_x)\in F_q$ record the number of the substructures $s_x\in \mathbf{S}$ appearing in $G_p$ and $G_q$, and $\mathbf{S}$ is the set of all possible substructures over all graphs in $\mathbf{G}$ based on $\mathrm{F}: G_p \rightarrow S_p$.

Generally speaking, within the scenario of R-convolution, one can adopt any graph decomposing approach to define a new R-convolution graph kernel, e.g., the R-convolution graph kernel based on the decomposed (a) path-based substructures~\cite{DBLP:journals/tnn/AzizWH13}, (b) walk-based substructures~\cite{DBLP:conf/sdm/KangTS12,DBLP:conf/nips/SugiyamaB15},  (c) subtree-based substructures~\cite{DBLP:journals/jmlr/AzaisI20,DBLP:journals/tcyb/Bai20}, (d) subgraph-based substructures~\cite{DBLP:conf/icml/KriegeM12,DBLP:journals/pr/BaiH16}, etc. For instance, Kashima et al.~\cite{DBLP:conf/icml/KashimaTI03}
have defined a Random-Walk Kernel based on the decomposed random walk based substructures. Borgwardt et al.~\cite{DBLP:conf/icdm/BorgwardtK05} have developed a Shortest-Path kernel based on the decomposed shortest path based substructures. Aziz et al.~\cite{DBLP:journals/tnn/AzizWH13} have proposed a Backtrackless-Path Kernel based on the non-backtrack cycle based substructures, that are identified by the Ihara zeta function~\cite{DBLP:journals/tnn/RenWH11}. Costa and Grave~\cite{DBLP:conf/icml/CostaG10} have defined a Pairwise-Distance Neighborhood-Subgraph Kernel based on decomposing graphs into layer-wise expansion neighborhood subgraphs, that rooted at pairwise vertices having a specified distance between them. Shervashidze et al.~\cite{shervashidze2010weisfeiler} have developed a Weisfeiler-Lehman Subtree Kernel based on the decomposed subtree based substructures, that are corresponded by the Weisfeiler-Lehman subtree invariants. Other prevalent R-convolution graph kernels include (a) the Pyramid Quantized Shortest-Path Kernel~\cite{DBLP:journals/ijon/GkirtzouB16}, (b) the Pyramid Quantized Weisfeiler-Lehman Subtree Kernel~\cite{DBLP:journals/ijon/GkirtzouB16}, (c) the Wasserstein Weisfeiler-Lehman Kernel~\cite{DBLP:conf/nips/TogninalliGLRB19}, (d) the Isolation Graph Kernel~\cite{DBLP:conf/aaai/XuTJ21}, (e) the Graph Filtration Kernel~\cite{DBLP:conf/aaai/SchulzWW22}, (f) the Attributed-Subgraph Matching Kernel~\cite{DBLP:conf/icml/KriegeM12}, (g) the Binary-Subtree Kernel~\cite{DBLP:conf/bmvc/GaidonHS11}, etc.

Unfortunately, there are three common drawbacks arising in most state-of-the-art R-convolution graph kernels. First, both Eq.(\ref{RK}) and Eq.(\ref{RK_PD}) indicate that the R-convolution kernel $\mathrm{K}_\mathrm{R}$ only focuses on comparing the isomorphic substructures shared by a pair of graphs, and thus neglect the influence of other unshared substructures between graphs, i.e., the unshared substructures do not participate the kernel computation. As a result, the R-convolution kernel $\mathrm{K}_\mathrm{R}$ may not reflect the precise similarity measures between graphs. For instance, Fig.\ref{ExampleG} exhibits four sample graphs as well as their associated feature vectors in terms of the substructures $\mathbf{S}=\{s_1,s_2,s_3,s_4,s_5\}$. Since the pair of graphs $G_1$ and $G_3$ and the pair of graphs $G_2$ and $G_3$ share the same number of isomorphic substructure pairs, the kernel values $\mathrm{K}_\mathrm{R}(G_1,G_2)$ and $\mathrm{K}_\mathrm{R}(G_2,G_3)$ between the two pairs of graphs are the same. Although, $G_1$ and $G_2$ are structurally different, and there are two unshared substructures $s_2$ and $s_3$ that have indicated the structural difference between them. The kernel values $\mathrm{K}_\mathrm{R}(G_1,G_2)$ and $\mathrm{K}_\mathrm{R}(G_2,G_3)$ still can not reflect the structural difference between $G_1$ and $G_2$. Second, the R-convolution kernels cannot reflect the distinctions between the isomorphic substructures in terms of the global graph structures. As a result, if a pair of graphs have the same numbers of all different substructures, the R-convolution kernels may not discriminate the intrinsic structural difference between the graphs. For instance, both the sample graphs $G_2$ and $G_4$ in Fig.\ref{ExampleG} have the same numbers of different substructures and the same feature vectors, thus the pair of graphs $G_2$ and $G_3$ and the pair of graphs $G_4$ and $G_3$ have the same kernel values. Although $G_2$ and $G_4$ are structurally different and their shared isomorphic substructures have different structural arrangement within the global graph structures. Third, the R-convolution kernels cannot reflect characteristics of global graph structures. This is due to the fact that the R-convolution kernels rely on the graph decomposition that may cause the notorious computational inefficiency. As a result, the R-convolution kernels tend to use substructures of small sizes, only reflecting local structural information. Overall, the above theoretical drawbacks significantly influence the performance of the R-convolution kernels. It is fair to say that developing effective graph kernels is always a theoretical challenge problem.

\subsection{Contributions of This Work}\label{s1.2}

In this work, we aim to overcome the theoretical drawbacks of the R-convolution graph kernels. To this end, we propose a novel quantum kernel, namely the Quantum-based Entropic Subtree Kernel (QESK), for graph classification problems. One key innovation of the proposed QESK kernel is to compute the entropic subtree representations for each graph through the Continuous-time Quantum Walk (CTQW)~\cite{DBLP:journals/pr/BaiH14B}. We show that the entropic subtree representations not only better discriminate the distinctions between isomorphic subtrees but also simultaneously reflect the global and local structural information through the CTQW. This in turn provides an elegant way to define new subtree-based kernels, and the resulting QESK kernel for a pair of graphs is defined by measuring the similarity between their associated entropic subtree representations. Overall, the contributions of this paper are summarized as follows.

\textbf{First}, for a set of graphs, we commence by evolving the CTQW on each graph structure and compute the associated Average Mixing Matrix (AMM), that encapsulates the visiting information of the CTQW. The reasons of employing the AMM matrix of the CTQW is that it not only reflects richer graph topological information in terms of the CTQW, but also assigns each vertex an individual probability distribution of the CTQW visiting all vertices and departing from the vertex (see details in Section~\ref{s2.2}). Thus, the AMM matrix can encapsulate complicated intrinsic structure information of graphs through the evolution of the CTQW. More specifically, we show how the AMM matrix allows us to compute the entropic subtree representations associated with the classical Weisfeiler-Lehman (WL) Tree-Index algorithm~\cite{UWL}. Unlike the original WL subtree representations, the proposed entropic subtree representations can not only reflect the structural arrangement difference between isomorphic subtrees rooted from different vertices, but also capture the global information of the whole graph structures. As a result, the entropic subtree representations can better represent the structural characteristics of graph structures.

% (see details in Section~\ref{s3.1})

\textbf{Second}, with a pair of graphs to hand, we define the QESK kernel by computing the exponentiation of the negative Euclidean distance between their entropic subtree representations, and this in turn results in a positive definite kernel. We show that the proposed QESK kernel can not only discriminate the graphs having the same numbers of different subtrees and encapsulate global graph structure information through the entropic subtree representations, but also integrate the effects of unshared subtrees into the kernel computation. As a result, the proposed QESK can reflect more precise kernel-based similarity measure than existing R-convolution kernels and significantly overcome the theoretical drawbacks of classical R-convolution kernels mentioned in Section~\ref{s1.1}, explaining the effectiveness of the proposed QESK kernel.

\textbf{Third}, we empirically investigate the classification performance of the proposed QESK kernel associated with the C-SVM on standard graph datasets. We demonstrate that the proposed QESK can outperform classical graph kernel and graph deep learning methods in terms of the classification accuracies.

\subsection{Paper Outline}

The remainder of this paper is organized as follows. Section \ref{s2} briefly reviews some related works that will be used in this work, including the classical WL Tree-Index algorithm and the concept of the CTQW. Section \ref{s3} gives the theoretical definition of the proposed QESK kernel. Section \ref{s4} empirically evaluates the performance of the proposed QESK kernel on graph classification tasks. Section \ref{s5} concludes of this work.

\section{Reviews of Preliminary Concepts}\label{s2}

In this section, we first review the WL Tree-Index algorithm~\cite{UWL}. Moreover, we show that how this algorithm can be adopted to define a Weisfeiler-Lehman Subtree Kernel (WLSK)~\cite{shervashidze2010weisfeiler}. Finally, we introduce the concept of the CTQW~\cite{DBLP:journals/pr/BaiH14B}.

\subsection{The Weisfeiler-Lehman Subtree-Index Method}\label{s2.1}

\begin{figure}
\centering
\subfigure{\includegraphics[width=1.0\linewidth]{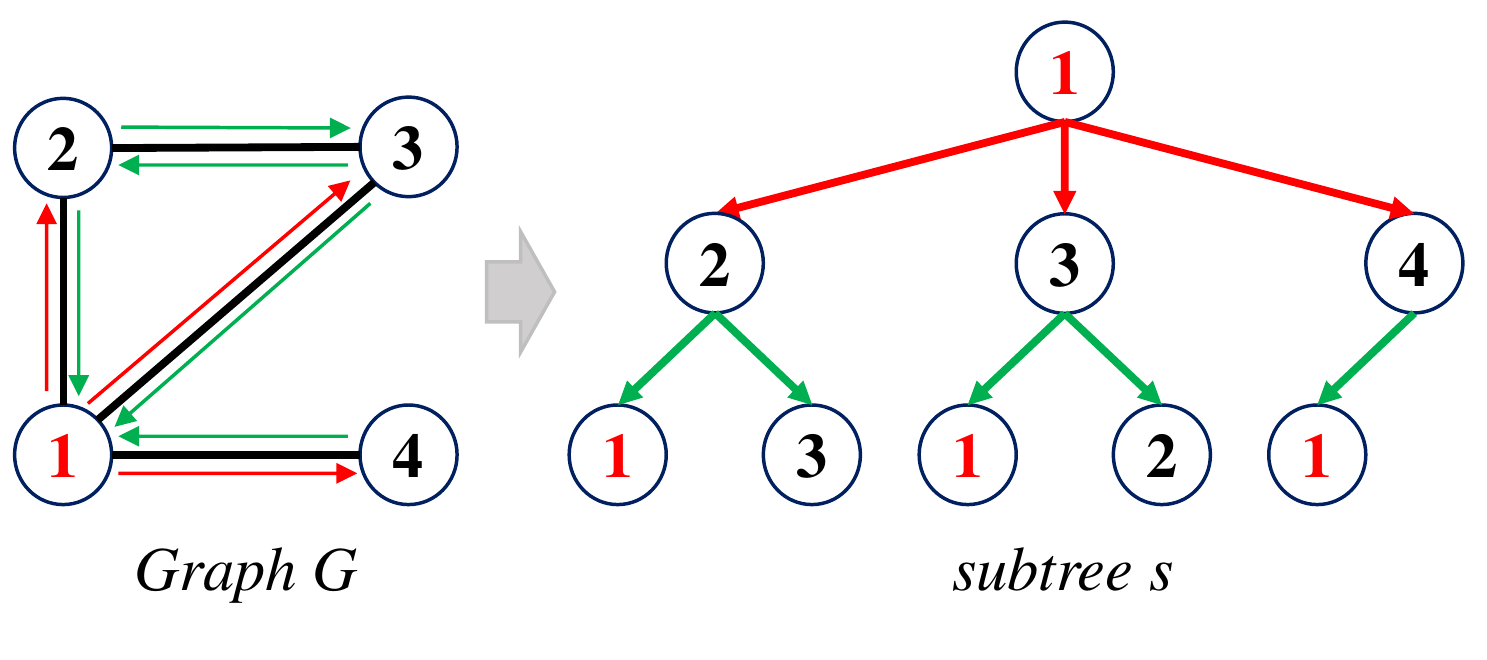}}
\vspace{-15pt}
\caption{Example for a subtree of height $2$ rooted at vertex $1$.}\label{ExampleS}
\vspace{-15pt}
\end{figure}

We introduce the WL Tree-Index method~\cite{UWL}, that can strengthen the vertex attribute information and further generate subtree invariants. Specifically, for the set of graphs $\mathbf{G}$, assume $G(V,E)\in \mathbf{G}$ is a sample graph with the vertex set $V$ and edge set $E$, $\mathrm{H}: \mathcal{L}_{\mathbf{G}} \rightarrow \Sigma$ is a Hush function that can convert any current vertex attribute of the graphs in $\mathbf{G}$ into a new integer-based attribute from the alphabet $\Sigma$, and $\mathcal{N}(v)=\{u|(v,u)\in E \}$ is the set of neighborhood vertices of $v$. The definition of the WL Tree-Index method is defined as follows:
\begin{enumerate}
  \item{Set the iteration number $I$ as $1$, i.e., $I=1$. For each vertex $v \in V$ and its original attribute $\mathcal{L}(v)$, employ the Hush function $\mathrm{H}: \mathcal{L}_{\mathbf{G}} \rightarrow \Sigma$ to transform $\mathcal{L}(v)$ into corresponding integers as the initialized attribute of $v$, i.e.,
      \begin{equation}
      \mathcal{L}^{I}(v)= \mathrm{H} [\mathcal{L}(v)].
      \end{equation}}
  \item {For each vertex $v \in V$, permute the attributes of its neighbourhood $\mathcal{N}(v)$ with the ascending order and construct an attribute list as
      \begin{equation}
      \mathcal{L}_{\mathcal{N}}^I(v)=\{\mathcal{L}^{I}(u)|u\in \mathcal{N}(v)\}.\label{owl1}
      \end{equation}
      }
  \item {Set $I=I+1$, strengthen the attribute $\mathcal{L}^{I}(v)$ of $v\in V$ by  taking the union of $\mathcal{L}^{I}(v)$ and $\mathcal{L}_{\mathcal{N}}^I(v)$ as
      \begin{equation}
      \mathcal{L}^{I}_{\mathrm{U}}(v)=\bigcup \{\mathcal{L}^{I-1}(v);\mathcal{L}_{\mathcal{N}}^{I-1}(v)\},\label{owl2}
      \end{equation}
      where $\mathcal{L}^{\mathrm{I}}_{U}(v)$ is now an attribute list.}
  \item {Convert the attribute list $\mathcal{L}^{I}(v)$ into a new integer-based attribute for $v\in V$ by using the Hash function $\mathrm{H}: \mathcal{L}_{\mathbf{G}} \rightarrow \Sigma$, i.e.,
      \begin{equation}
      \mathcal{L}^{I}(v)=\mathrm{H} [\mathcal{L}^{I}_{\mathrm{U}}(v)].\label{owl3}
      \end{equation}}
  \item {Check the value of iteration $I$. If $I$ achieves the predetermined value, terminate the operation and output the current attributes of all vertices in $V$. Otherwise, repeat the computational procedures from $2$ to $5$.}
\end{enumerate}

\begin{figure*}
\centering
\subfigure{\includegraphics[width=1.0\linewidth]{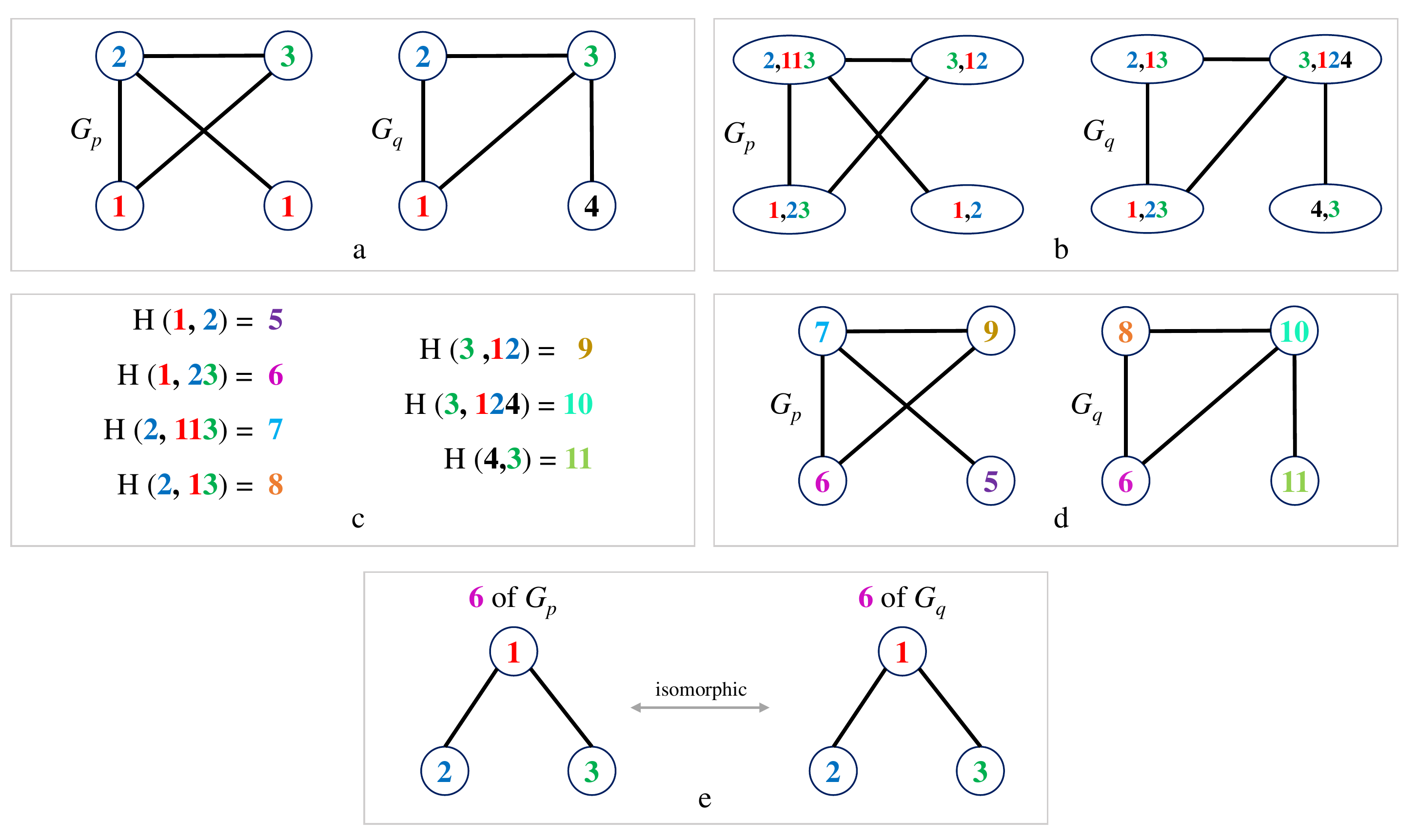}}
\vspace{-25pt}
\caption{The procedures of identifying isomorphic subtree-based substructures between graphs $G_p(V_p,E_p)\in \mathbf{G}$ and $G_q(V_q,E_q)\in \mathbf{G}$, based on the WL Tree-Index method. (a) Convert the original vertex attributes into the new integer-based attributes as the initialized vertex attributes based on the first step of the WL Tree-Index method. (b) Strengthen the attribute for each vertex based on the second and third steps of the WL Tree-Index method. (c) Generate new attribute for each vertex based on the fourth step of the WL Tree-Index method. (d) Identify the isomorphic subtrees between $G_p$ and $G_q$ based on their identical vertex attributes.} \label{ExampleWL}
\vspace{-15pt}
\end{figure*}

Note that, for the WL Tree-Index method, we always employ the same Hush function $H$ in the above computational procedures. This in turn guarantees that the identical strengthened vertex attributes over all graphs in $\mathbf{G}$ can be converted into the same corresponding integer-based attributes. For each iteration $I$ ($I\geq 1$), the attribute $\mathcal{L}^{I}(v)$ of vertex $v$ corresponds to a subtree of height $I$ rooted at $v$. Figure~\ref{ExampleS} exhibits the structure of a subtree with height $2$ rooted at vertex $1$. For a pair of graphs $G_p(V_p,E_p)\in \mathbf{G}$ and $G_q(V_q,E_q)\in \mathbf{G}$, if their attributes $\mathcal{L}^{I}(v_p)$ and $\mathcal{L}^{I}(v_q)$ are identical, the subtrees of height $I$ rooted at $v_p\in V_p$ and $v_q\in V_q$ are isomorphic to each other. Figure~\ref{ExampleWL} gives an detailed example to explain how to identify isomorphic subtree-based substructures based on the WL Tree-Index method. More specifically, Shervashidze et al.~\cite{DBLP:journals/jmlr/ShervashidzeSLMB11} have utilized the classical WL Tree-Index method to propose a Weisfeiler-Lehman Subtree Kernel (WLSK), by counting the numbers of identical vertex attribute pairs between pairwise of graphs. Although, the WLSK kernel can significantly outperform most R-convolution graph kernels in terms of the graph classification, the WLSK still suffers from the three aforementioned theoretical drawbacks arising in R-convolution kernels, influencing the further effectiveness improvement.

\subsection{The Continuous-time Quantum Walk}\label{s2.2}

One objective of this work is to utilize the CTQW to compute the entropic subtree representations, enhancing the capability of the structural representation. The reason of utilizing the CTQW is that the behaviour of the CTQW theoretically differs from its classical counterpart (i.e., the Continuous-time Random Walk (CTRW)~\cite{DBLP:journals/jcss/Watrous01}) and have a number of interesting properties. First, unlike the classical CTRW, the state vector of the CTQW is complex-valued rather than real-valued, and its evolution is also governed by an unitary matrix rather than a stochastic matrix. Thus, the CTQW evolution is reversible, indicating that the CTQW is non-ergodic, and not restricted by a limiting distribution. As a result, the CTQW can naturally reducing the notorious shortcoming of tottering arising in the classical CTRW, better reflecting structural information than the CTRW. Second, the CTQW evolution is not determined by the Laplacian spectrum with low frequency components. As a result, the CTQW can better discriminate different graph structures than the classical CTRW.

In this subsection, we introduce the concept of the CTQW as well as its associated AMM matrix. Specifically, for the sample graph $G(V,E)$, the state space of the CTQW is the vertex set $V$. Through the Dirac notation, the basis state of the CTQW at vertex $v \in V$ is written as $\Ket{v}$. Here, $\Ket{.}$ is the orthonormal vector represented in a $|V|$-dimensional complex-valued Hilbert space. Its state $\Ket{\psi(t)}$ at time $t$ is a complex linear combination of these orthonormal state vectors $\Ket{v}$, i.e.,
\begin{equation}
\Ket{\psi(t)} = \sum_{v\in V} \alpha_v(t) \Ket{v},
\end{equation}
where $\alpha_v (t) \in \mathbb{C}$ is the complex amplitude. Unlike the classical counterpart, the evolution of the CTQW is based on the Schr\"{o}dinger equation, i.e.,
\begin{equation}
\frac{\partial \Ket{\psi_t}}{\partial t}  = -i\mathcal{H}\Ket{\psi_t},
\end{equation}
where $\mathcal{H}$ represents the system Hamiltonian and accounts for the total energy of the system, and one can adopt the adjacency matrix as the Hamiltonian.

Specifically, the behaviour of the CTQW over $G(V,E)$ at time $t$ can be described with the AMM matrix~\cite{godsil2013average}, i.e.,
\begin{align}\label{}
Q_\mathrm{M}(t) &= U(t) \circ U(-t) \nonumber \\
&= e^{i\mathcal{H}t} \circ e^{-i\mathcal{H}t},
\end{align}
where the symbol $\circ$ represents the operation of the Schur-Hadamard product between $e^{i\mathcal{H}t}$ and $e^{-i\mathcal{H}t}$. Note that, since $U$ is unitary, $Q_\mathrm{M}(t)$ is a doubly stochastic matrix and its entry $Q_\mathrm{M}(t)_{uv}$ corresponds to the probability of the CTQW visiting vertex $v\in V$ at time $t$ when the CTQW departs from $u\in V$. To guarantee the convergence of $Q_\mathrm{M}(t)$, we can take the Ces\`{a}ro mean and compute a time-average based AMM matrix of the CTQW as
\begin{equation}
Q = \lim_{T \rightarrow \infty} \int_{0}^{T} Q_M(t) dt,
\end{equation}
where each entry $Q_{vu}$ of the AMM matrix $Q$ corresponds to the average probability of the CTQW visiting $u\in V$ and departing from $v\in V$, and $Q$ is still a doubly stochastic matrix.

Furthermore, since Godsil~\cite{godsil2013average} has indicated that the entries of $Q$ are rational numbers, thus one can easily compute $Q$ from the spectrum of the Hamiltonian. Specifically, let the adjacency matrix $A$ of $G$ be the Hamiltonian $\mathcal{H}$, $\lambda_1,\ldots,\lambda_{|V|}$ represent the $|V|$ distinct eigenvalues of $H$, and $\mathbb{P}_j$ be the matrix representation of the orthogonal projection on the eigenspace associated with the $\lambda_j$ (i.e., $\mathcal{H} = \sum_{j = 1}^{|V|} \lambda_j \mathbb{P}_j$), the AMM matrix $Q$ of the CTQW can be rewritten as
\begin{equation}
Q = \sum_{j = 1}^{|V|} \mathbb{P}_j \circ \mathbb{P}_j.
\end{equation}

\section{The Quantum-based Entropic Subtree Kernel}\label{s3}

We define a novel Quantum-based Entropic Subtree Kernel (QESK) for attributed graphs. We first propose the concept of entropic subtree representations. Moreover, we define the QESK kernel between graphs. Finally, we discuss the theoretical advantages of the proposed QESK kernel.

\subsection{Quantum-based Entropic Subtree Representations}\label{s3.1}

In this subsection, we propose a new framework of computing the entropic subtree presentations to represent the structural characteristics of graphs, associated with the CTQW as well as the classical WL Tree-Index method. Specifically, for the sample graph $G(V,E)$ from the graphs set $\mathbf{G}$, we commence by computing the AMM matrix $Q$ of the CTQW evolved on $G$. Moreover, we perform the classical WL Tree-Index method on $G$ by varying the iteration number $I$ from $1$ to $I_{\mathrm{max}}$ (i.e., $1\leq I \leq I_{\mathrm{max}}$), and obtain $I_{\mathrm{max}}$ new attributes for each vertex $v\in V$. For each vertex $v\in V$, since the $v$-th row or collum of the AMM matrix $Q$ corresponds to the time-averaged probability distribution of the CTQW visiting all vertices $u\in V$ when the CTQW starts form vertex $v\in V$. We propose a quantum-inspired Shannon entropy $H_{\mathrm{QS}}(v)$ for each vertex $v\in V$ associated with the $v$-th row of $Q$, i.e.,
\begin{equation}
H_\mathrm{QS}(v)=-\sum_{u\in V} Q_{vu} \log Q_{vu}.
\end{equation}

With the quantum-inspired Shannon entropies of all vertices in $G(V,E)$ to hand, we compute the entropic subtree representation associated with the vertex attributes. Assume $\mathbf{L}_\mathbf{G}^I=\{\mathcal{L}_1^I, \ldots,\mathcal{L}_x^I, \ldots,\mathcal{L}_{M_I}^I \}$ ($|\mathbf{L}_\mathbf{G}^I|=M_I$) is the set of all possible $M_I$ vertex attributes for the set of vertices $\mathbf{V}$ over all graphs in $\mathbf{G}$, through the $I$-th iteration of the WL Tree-Index method. For each graph $G(V,E)\in \mathbf{G}$, its $I$-level entropic subtree representation $F_{\mathrm{QS}}^I(G)$ is defined as
\begin{equation}
F_{\mathrm{QS}}^I(G)=\{\mathrm{E}(\mathcal{L}_1^I),\ldots, \mathrm{E}(\mathcal{L}_x^I),\ldots,\mathrm{E}(\mathcal{L}_{M_I}^I)  \},
\end{equation}
where each element $\mathrm{E}(\mathcal{L}_x^I)$ is the entropic subtree pattern of the subtree corresponded by $\mathcal{L}_x^I$, and is defined as
\begin{equation}
\mathrm{E}(\mathcal{L}_x^I)={ \sum_{v\in V\{\mathcal{L}_x^I\}}H_\mathrm{QS}(v)   }/{ \sum_{v\in V}H_\mathrm{QS}(v) }.\label{Entropic_subtree}
\end{equation}
Here, $V\{\mathcal{L}_x^I\}$ is the set of vertices having the same vertex attribute $\mathcal{L}_x^I$, and $V\{\mathcal{L}_x^I\}\in V$. Clearly, the $I$-level entropic subtree representation $F_{\mathrm{QS}}^I(G)$ can be seen as an entropic distribution of the quantum-inspired Shannon entropies over all vertices in $V$ of $G$, and is spanned by the specific subtree invariants that are corresponded by the vertex attributes of $\mathbf{L}_\mathbf{G}^I$.

\vspace{5pt}

\noindent\textbf{Remarks:} Note that, for the AMM matrix $Q$ of the CTQW, the different $v$-th rows are usually different for complicated graph structures, indicating that the time-averaged probability distributions of the CTQW visiting all vertices of $G$ are distinct if the CTQW departs from different vertices $v\in V$. In other words, these probability distributions of the CTQW over all vertices can reflect multi-viewed global intrinsic graph structure information in terms of different local starting vertices. As a result, the entropic subtree representations can simultaneously capture the global and local structural information of graphs, through the AMM matrix $Q$.

% i.e., each local vertex corresponds to a specific global probability distributions of the CTQW.

%\begin{equation}
%F_{\mathrm{QS}}^I(G)=\{ \frac{ \sum_{v\in V\{\mathcal{L}_1^I\}}H_\mathrm{QS}(v)   }{ \sum_{v\in V}H_\mathrm{QS}(v) },\ldots,\frac{ \sum_{v\in V\{\mathcal{L}_x^I\}}H_\mathrm{QS}(v)   }{ \sum_{v\in V}H_\mathrm{QS}(v) },\ldots,\frac{ \sum_{v\in V\{\mathcal{L}_\mathbf{N}^I\}}H_\mathrm{QS}(v)   }{ \sum_{v\in V}H_\mathrm{QS}(v) }  \},
%\end{equation}

\subsection{The Definition of the Proposed QESK Kernel}\label{s3.2}
In this subsection, we define a novel QESK kernel between a pair of graphs associated with their entropic subtree representations defined in Section~\ref{s3.2}.

\vspace{5pt}

\noindent \textbf{Definition (The Quantum-based Entropic Subtree Kernel)}: For the pair of graphs $G_p\in \mathbf{G}$ and $G_q\in \mathbf{G}$ defined previously, assume
$$F_{\mathrm{QS}}^I(G_p)=\{\mathrm{E}_p(\mathcal{L}_{1}^I),\ldots, \mathrm{E}_p(\mathcal{L}_{x}^I),\ldots,\mathrm{E}_p(\mathcal{L}_{M_I}^I)  \}$$
and
$$F_{\mathrm{QS}}^I(G_q)=\{\mathrm{E}_p(\mathcal{L}_{1}^I),\ldots, \mathrm{E}_p(\mathcal{L}_{x}^I),\ldots,\mathrm{E}_p(\mathcal{L}_{M_I}^I)  \}$$
are their associated $I$-level entropic subtree representations computed from the $I$-th iteration of the WL Tree-Index method defined in Section~\ref{s2.1}. The proposed QESK kernel $\mathrm{K}_{\mathrm{QESK}}$ between $G_p$ and $G_q$ is defined by computing the exponentiation of
the negative Euclidean distance between their entropic subtree representations as
\begin{align}
\mathrm{K}_{\mathrm{QESK}}(G_p,G_q)&=\sum_{I=1}^{I_\mathrm{max}} \mathrm{K}_{\mathrm{QESK}}^I(G_p,G_q) \nonumber \\
&=\sum_{I=1}^{I_\mathrm{max}} \exp  (-\| F_{\mathrm{QS}}^I(G_p)- F_{\mathrm{QS}}^I(G_p) \|) \nonumber \\
&= \sum_{I=1}^{I_\mathrm{max}}
\exp\{-\sqrt{\sum_{x=1}^{M_I} [\mathrm{E}_p(\mathcal{L}_{x}^I)-\mathrm{E}_q(\mathcal{L}_{x}^I)]^2} \},\label{QESK}
\end{align}
where $I_\mathrm{max}$ is the greatest number of the iteration parameter $I$ for the WL Tree-Index method, and each $\mathrm{K}_{\mathrm{QESK}}^I$ is the $I$-level QESK kernel defined as
\begin{align}
\mathrm{K}_{\mathrm{QESK}}^I(G_p,G_q)&=
\exp\{-\sqrt{\sum_{x=1}^{M_I} [\mathrm{E}_p(\mathcal{L}_{x}^I)-\mathrm{E}_q(\mathcal{L}_{x}^I)]^2} \}.\label{QESK_I}
\end{align}
In this work, we propose to set $I_\mathrm{max}=10$, indicating that the proposed QESK kernel $\mathrm{K}_{\mathrm{QESK}}$ usually focuses on the subtree structures whose heights are equal or lower than $10$. \hfill$\blacksquare$

\vspace{5pt}

\noindent\textbf{Lemma.} \emph{The QESK kernel is positive definite (\textbf{pd}).}

\vspace{5pt}

\noindent\textbf{Proof.} This follows the fact that a diffusion kernel $\mathrm{K}_\mathrm{Diff}=\exp(-\lambda \mathrm{D}(G_p,G_q))$ associated with a dissimilarity or distance measure $\mathrm{D}$ between $G_p$ and $G_q$ is \textbf{pd}, if $\mathrm{D}_G$ is symmetrical. Since the Euclidean distance measure is symmetric, the $I$-level QESK kernel is \textbf{pd}. As a result, the proposed QESK kernel can be seen as the sum of several \textbf{pd} kernels, and is also \textbf{pd}.

\vspace{5pt}

\noindent\textbf{Remarks:} Since the proposed QESK kernel is defined based on the entropic subtree representations, where each element corresponds to an entropic pattern of a specific subtree invariant rather than the number of the subtree invariant appearing in the graph. The proposed QESK kernel can reflect richer quantum entropy-based structural similarity information than the classical R-convolution graph kernels, that only focus on counting the number of isomorphic substructure pairs. In Section~\ref{s3.3}, we will further analyze the theoretical advantages of the proposed QESK kernel, explaining the effectiveness.

\subsection{Advantages of the Proposed QESK Kernel}\label{s3.3}

%\begin{table*}
%\centering {
%% \tiny
%%\scriptsize
%\footnotesize
%\caption{Properties of the Proposed HAQJSK Kernels}\label{Comparison}
%% \scriptsize
%\vspace{-0pt}
%\begin{tabular}{|c||c||c||c||c||c|}
%
%  \hline
%  % after \\: \hline or \cline{col1-col2} \cline{col3-col4} ...
% ~Kernel Properties ~ & ~\textbf{QESK}  ~      &~JTQK~    & ~QWGK~ &  ~RCGK~ &  ~GGK~\\ \hline \hline
%
% ~Capture Local Structural Information~             & ~$\mathrm{Yes}$~  &~ $\mathrm{Yes}$~  & ~$\mathrm{Yes}$~&  ~$\mathrm{Yes}$~&  ~$\mathrm{No}$~ \\  \hline
%
% ~Capture Global Structural Information~            & ~$\mathrm{Yes}$~  &~ $\mathrm{Yes}$~  & ~$\mathrm{No}$~ &  ~$\mathrm{No}$~ &  ~$\mathrm{Yes}$~ \\  \hline
%
% ~Accommodate Vertex Attributes~              & ~$\mathrm{Yes}$~  &~ $\mathrm{No}$~   & ~$\mathrm{No}$~ &  ~$\mathrm{-}$~ &  ~$\mathrm{-}$~ \\  \hline
%
% ~Accommodate Unshared Substructures~              & ~$\mathrm{Yes}$~  &~ $\mathrm{No}$~   & ~$\mathrm{No}$~ &  ~$\mathrm{-}$~ &  ~$\mathrm{-}$~ \\  \hline
%
% ~Discriminate Isomorphic Substructures~        & ~$\mathrm{Yes}$~  &~ $\mathrm{No}$~   & ~$\mathrm{No}$~ &  ~$\mathrm{-}$~ &  ~$\mathrm{-}$~ \\  \hline
%
%\end{tabular}
%}
%\begin{tablenotes}
%\item[1] $-$: indicate that these kernels do not refer to this problem.
%\end{tablenotes}
%\vspace{-15pt}
%\end{table*}

In this subsection, we indicate the advantages of the proposed QESK kernel by revealing the theoretical difference between the QESK kernel and the state-of-the-art R-convolution graph kernels. As we have stated in Section~\ref{s1.2}, most R-convolution graph kernels usually suffer from three common theoretical drawbacks, i.e., (a) neglect the effects of unshared substructures between graphs, (b) can not reflect the distinctions between isomorphic substructures in terms of the structural arrangement of global graph structures, and (c) only focus on structural information of local substructures. To further indicate how the proposed QESK kernel addresses the three drawbacks, we first give the definition of the WLSK kernel that have been mentioned in Section~\ref{s2.1} and is a typical instance of the R-convolution kernels. For the pair of graphs $G_p\in \mathbf{G}$ and $G_q\in \mathbf{G}$ defined previously, assume
$$F_{\mathrm{WL}}^I(G_p)=\{\mathrm{N}_p(\mathcal{L}_{1}^I),\ldots, \mathrm{N}_p(\mathcal{L}_{x}^I),\ldots,\mathrm{N}_p(\mathcal{L}_{M_I}^I)  \}$$
and
$$F_{\mathrm{WL                                       }}^I(G_q)=\{\mathrm{N}_q(\mathcal{L}_{1}^I),\ldots, \mathrm{N}_q(\mathcal{L}_{x}^I),\ldots,\mathrm{N}_q(\mathcal{L}_{M_I}^I)  \}$$
are their associated subtree invariant vectors, where each elements $\mathrm{N}_p(\mathcal{L}_{x}^I)$ and $\mathrm{N}_q(\mathcal{L}_{x}^I)$ record the numbers of the subtree invariants corresponded by $\mathcal{L}_{x}^I\in \mathbf{L}_\mathbf{G}^I$ and appearing in $G_p$ and $G_q$. The WLSK kernel $\mathrm{K}_{\mathrm{WLSK}}$ between $G_p$ and $G_q$ is computed by counting the numbers of shared isomorphic subtrees~\cite{shervashidze2010weisfeiler}, and is defined as
\begin{align}
\mathrm{K}_{\mathrm{QESK}}(G_p,G_q)&=\sum_{I=1}^{I_\mathrm{max}} \mathrm{K}_{\mathrm{WLSK}}^I(G_p,G_q) \nonumber \\
&=\sum_{I=1}^{I_\mathrm{max}} \langle F_{\mathrm{WL}}^I(G_p), F_{\mathrm{WL}}^I(G_q) \rangle  \nonumber \\
&= \sum_{I=1}^{I_\mathrm{max}} \sum_{x=1}^{M_{I}} \mathrm{N}_p(\mathcal{L}_{x}^I) \mathrm{N}_q(\mathcal{L}_{x}^I),\label{WLSK}
\end{align}
where each $\mathrm{K}_{\mathrm{WLSK}}^I$ is the $I$-level WLSK kernel defined as
\begin{align}
\mathrm{K}_{\mathrm{QESK}}(G_p,G_q)= \sum_{x=1}^{M_{I}} \mathrm{N}_p(\mathcal{L}_{x}^I) \mathrm{N}_q(\mathcal{L}_{x}^I).\label{WLSK_I}
\end{align}
Specifically, Eq.(\ref{QESK}) and Eq.(\ref{QESK_I}) of the proposed QESK kernel, and Eq.(\ref{WLSK}) and Eq.(\ref{WLSK_I}) of the WLKS kernel exhibit the following theoretical differences between the two kernels.

\textbf{(a)} For the pair of graphs $G_p$ and $G_q$, the proposed QESK kernel relies on computing the distance between their entropic subtree patterns $\mathrm{E}_p(\mathcal{L}_{x}^I)$ and $\mathrm{E}_q(\mathcal{L}_{x}^I)$ of the subtree corresponded by $\mathcal{L}_{x}^I$. Since the distance value between $\mathrm{E}_p(\mathcal{L}_{x}^I)$ and $\mathrm{E}_q(\mathcal{L}_{x}^I)$ will not be zero, when the subtree only exists in one of $G_p$ and $G_q$, i.e., the subtree is not shared between $G_p$ and $G_q$. As a result, the unshared subtree can still influence the kernel computation, and the proposed QESK kernel is able to reflect the effect of unshared subtrees between graphs. By contrast, the classical WLSK kernel relies on computing the product between the numbers $\mathrm{N}_p(\mathcal{L}_{x}^I)$ and $\mathrm{N}_q(\mathcal{L}_{x}^I)$ of the subtree corresponded by $\mathcal{L}_{x}^I$ and appearing in $G_p$ and $G_q$, and the product value will be zero if this subtree only exists in one of $G_p$ and $G_q$. As a result, the classical WLSK kernel fails to reflect the effect of the unshared subtree between $G_p$ and $G_p$.

On the other hand, for the proposed QESK kernel, Eq.(\ref{Entropic_subtree}) indicates that its entropic subtree patterns $\mathrm{E}_p(\mathcal{L}_{x}^I)\in F_{\mathrm{QS}}^I(G_p)$ and $\mathrm{E}_q(\mathcal{L}_{x}^I)\in F_{\mathrm{QS}}^I(G_q)$ are computed associated with the quantum-inspired Shannon entropies over all vertices, including the vertices with unshared vertex attributes $\mathcal{L}_{x}^I$ between $G_p$ and $G_q$ (i.e., the unshared subtree corresponded by $\mathcal{L}_{x}^I$). Thus, the resulting entropic subtree representations $F_{\mathrm{QS}}^I(G_p)$ and $F_{\mathrm{QS}}^I(G_q)$ can potentially reflect the effect of the unshared subtrees for the QESK kernel between $G_p$ and $G_q$. By contrast, for the classical WLSK kernels, the elements $\mathrm{N}_p(\mathcal{L}_{x}^I)$ and $\mathrm{N}_q(\mathcal{L}_{x}^I)$ of its subtree invariant vectors $F_{\mathrm{WL}}^I(G_p)$ and $F_{\mathrm{WL}}^I(G_q)$ only record the numbers of a specific subtree appearing in $G_p$ and $G_q$, and cannot directly reflect any information in terms of the unshared subtrees between $G_p$ and $G_q$.

\textbf{(b)} For the classical WLSK kernel, if the pair of graphs $G_p$ and $G_q$ are structurally different but both have the same numbers of different isomorphic subtrees corresponded over all different $\mathcal{L}_{x}^I$ that exist in $G_p$ and $G_q$, their subtree invariant vectors $F_{\mathrm{WL}}^I(G_p)$ and $F_{\mathrm{WL}}^I(G_q)$ will be the same. The WLSK kernel can not discriminate the structural difference between such graphs, since it cannot identify any structural distinction between isomorphic subtrees through their corresponded vertex attributes that are identical. Although the vertices having the same attribute may have different structural arrangements within their global graph structures.  By contrast, for the proposed QESK kernel, since the rows of the AMM matrix $Q$ are usually different for complicated graph structures, we may still assign different time-averaged probability distributions of the CTQW to the vertices that have the same vertex attribute $\mathcal{L}_{x}^I$. In other words, the AMM matrix $Q$ can significantly reflect structural distinctions between isomorphic subtrees corresponded by $\mathcal{L}_{x}^I$, in terms of the complicated arrangement of global graph structures. As a result, the entropic subtree representations $F_{\mathrm{QS}}^I(G_p)$ and $F_{\mathrm{QS}}^I(G_p)$ of $G_p$ and $G_q$ can still be different, and the proposed QESK kernel is able to discriminate $G_p$ and $G_q$.

\textbf{(c)} As we have stated in Section~\ref{s3.1}, the associated entropic subtree representations of the proposed WLSK kernel can simultaneous capture the global and local graph structure information through the AMM matrix of the CTQW. As a result, the proposed QESK kernel can also reflect either the global or local structural characteristics of graph structures associated with the entropic subtree representations. By contrast, the classical WLSK focuses on counting the number isomorphic subtrees can only reflect local structural information.

The above theoretical analysis indicates that the proposed QESK kernel can capture more precise kernel-based similarity measures between graphs than R-convolution graph kernels.

\subsection{Computational Complexity}\label{s3.4}

For the graph set $\mathbf{G}$, assume there are $N$ graphs in $\mathbf{G}$ and each graph $G\in \mathbf{G}$ has $n$ vertices. Computing the proposed QESK kernel $\mathrm{K}_{\mathrm{QESK}}$ over all the $N^2$ pairs of graphs from $\mathbf{G}$ relies on the following computational steps. \textbf{(a)} Computing the vertex attribute set $\mathbf{L}_\mathbf{G}^I$ for each iteration $I$ of the WL Tree-Index method needs to consider all the $n^2$ elements of the adjacency matrix for each graph. For the $N$ graphs in $\mathbf{G}$, this process thus requires time complexity $O(I_{\mathrm{max}}Nn^2)$ where $I_{\mathrm{max}}$ is the greatest number of $I$. \textbf{(b)} Computing the entropic subtree representation relies on evolving the CTQW on each graph, that refers to the eigen-decomposition of the Laplacian. For the $N$ graphs in $\mathbf{G}$, this process thus requires time complexity $O(Nn^3)$. \textbf{(c)} Computing each $I$-leve QESK kernel $\mathrm{K}_{\mathrm{QESK}}^I$ between all pairs of graphs in $\mathbf{G}$ associated with the entropic subtree representations at most requires time complexity $O(N^2n)$, when the vertex attributes of a graph in $\mathbf{G}$ are all different.

As a result, computing the QESK kernel $\mathrm{K}_{\mathrm{QESK}}$ over $\mathbf{G}$ requires time complexity $O(I_{\mathrm{max}}Nn^2 + Nn^3 + I_{\mathrm{max}}N^2n)$. Since $I_{\mathrm{max}}\ll N$, the time complexity can be rewritten as $$O(Nn^3+N^2n).$$ For a pair of graphs ($N=2$), the time complexity should be $$O(n^3).$$ The computational analysis indicates that the proposed QESK kernel $\mathrm{K}_{\mathrm{QESK}}$ can be computed in a polynomial time.

\section{Experiments}\label{s4}

We compare the classification performance of the proposed QESK kernel with either graph kernels or graph deep learning approaches on nine benchmark graph datasets.

\subsection{Benchmark Datasets}
We employ nine benchmark graph datasets extracted from bioinformatics (BIO) and social networks (SN) for the performance evaluation. The BIO and SN datasets can be found in~\cite{KKMMN2016}, and the statistical information of these datasets is shown in Table~\ref{T:GraphInformation}.

\begin{table*}
\centering {
% \tiny
% \scriptsize
 \footnotesize
\vspace{-0pt}
\caption{Information of the Graph Datasets}\label{T:GraphInformation}
\vspace{-0pt}
\begin{tabular}{|c|c|c|c|c|c|c|c|c|c|}

  \hline
 ~Datasets ~          & ~MUTAG  ~ & ~NCI1~     & ~PROTEINS~& ~D\&D~       & ~PTC(MR)~  &  ~COLLAB  ~ & ~IMDB-B~   & ~IMDB-M~  & ~RED-B~ \\ \hline \hline

  ~Max \# vertices~   & ~$28$~    & ~$111$~    & ~$620$~   &  ~$5748$~    & ~$109$~    & ~$492$~    & ~$136$~     & ~$89$~    & ~$3782$~\\ \hline

  ~Mean \# vertices~  & ~$17.93$~ & ~$29.87$~  & ~$39.06$~ &  ~$284.30$~  & ~$25.56$~  & ~$74.49$~  & ~$19.77$~   & ~$13.00$~ & ~$429.62$~\\  \hline

  ~Mean \# edges~     & ~$19.79$~ & ~$32.30$~  & ~$72.82$~ &  ~$715.65$~  & ~$25.96$~  & ~$2457.50$~& ~$96.53$~  & ~$65.93$~& ~$497.75$~\\  \hline

  ~\# graphs~         & ~$188$~   &  ~$4110$~  & ~$1113$~  &  ~$1178$~    & ~$344$~    & ~$5000$~   &  ~$1000$~   & ~$1500$~  & ~$2000$~    \\ \hline

~\# vertex labels~    & ~$7$~     &  ~$37$~    & ~$61$~    &  ~$82$~      & ~$19$~     & ~$-$~      &  ~$-$~      & ~$-$~     & ~$-$~   \\ \hline

~\# classes~          & ~$2$~     &  ~$2$~     & ~$2$~     &  ~$2$~       & ~$2$~      &  ~$3$~     & ~$2$~       &  ~$3$~    & ~$2$~   \\ \hline

~Description~         & ~BIO~&  ~BIO~& ~BIO~&  ~BIO~  & ~BIO~ & ~Social~   &  ~Social~   & ~Social~  & ~Social~   \\ \hline

\end{tabular}

} \vspace{-0pt}
\end{table*}

\subsection{Experimental Comparisons with Graph Kernels}
\textbf{Experimental Setups:} We compare the performance of the proposed QESK kernel with some classical graph kernels, including (1) the Quantum Jensen-Shannon Kernel (QJSK) associated with the CTQW~\cite{DBLP:journals/pr/BaiH14B}, (2) the Jensen-Tsallis q-difference Kernel (JTQK)~\cite{DBLP:conf/pkdd/Bai0BH14} with $q=2$ and the subtrees of height $10$, (3) the Graphlet Count Graph Kernel (GCGK)~\cite{DBLP:journals/jmlr/ShervashidzeVPMB09} with graphlet of size $4$, (4) the classical Weisfeiler-Lehman Subtree Kernel (WLSK)~\cite{DBLP:journals/jmlr/ShervashidzeSLMB11} with the subtree  height $10$, (5) the WLSK kernel associated with Core-Variants (CORE WL)~\cite{DBLP:conf/ijcai/NikolentzosMLV18}, (7) the Shortest Path Graph Kernel (SPGK)~\cite{DBLP:conf/icdm/BorgwardtK05}, (7) the SPGK kernel associated with Core-Variants (CORE SP)~\cite{DBLP:conf/ijcai/NikolentzosMLV18}, (8) the Pyramid Match Graph Kernel (PMGK)~\cite{DBLP:conf/aaai/NikolentzosMV17}, (9) the PMGK kernel associated with Core-Variants (CORE PM)~\cite{DBLP:conf/ijcai/NikolentzosMLV18}, and (10) the Random-Walk Graph Kernel (RWGK)~\cite{DBLP:conf/icml/KashimaTI03}. Properties of all these kernels have been summarized in Table~\ref{T:GKInfor}. Specifically, we perform the $10$-fold cross-validation of the C-SVM~\cite{ChangLinSVM2001} associated with these graph kernels to compute the classification accuracies, using nine folds for training and one fold for testing.
For each kernel on each dataset, the parameters of the C-SVM are optimized based on the training set, and the experiments are repeated the for ten times. Table~\ref{T:ClassificationGK} shows the average classification accuracy ($\pm$ standard error). Note that, for some alternative kernels, the experimental results are directly from the original references or the comprehensive review paper~\cite{DBLP:journals/entropy/ZhangWW18}, following the same experimental setups with ours.

\begin{table*}
\centering {
%\tiny
% \scriptsize
 \footnotesize
\caption{Property Comparisons for Graph Kernels.}\label{T:GKInfor}
\vspace{-0pt}
\begin{tabular}{|c|c|c|c|c|c|c|}

  \hline
 ~Kernel Methods                                     ~& ~Kernel Frameworks~   & ~Global Kernels~      & ~Local Kernels~     & ~Discriminate~ & ~The Unshared~  & ~The Vertex~\\
 ~                                                   ~& ~                 ~   & ~              ~      & ~             ~     & ~Isomorphism~  & ~Substructures~ & ~Attributes~ \\ \hline \hline

  ~\textbf{QESK}~                                     & ~R-convolution~       & ~Yes (Multi-Viewed)~  &  ~Yes (Subtrees)~   & ~Yes~          & ~Accommodate~   & ~Accommodate~  \\ \hline

  ~QJSK~\cite{DBLP:journals/pr/BaiH14B}~              & ~Information Theoretic~  & ~Yes (Single-Viewed)~ &  ~No~               & ~- ~           & ~-~             & ~Neglect~\\ \hline

  ~JTQK~\cite{DBLP:conf/pkdd/Bai0BH14}~               & ~R-convolution~       & ~No~ &  ~Yes (Subtrees)~   & ~No~           & ~Neglect~       & ~Accommodate~ \\ \hline

  ~GCGK~\cite{DBLP:journals/jmlr/ShervashidzeVPMB09}~ & ~R-convolution~       & ~No~                  &  ~Yes (Graphlets)~  & ~No~           & ~Neglect~       & ~Neglect~\\ \hline

  ~WLSK~\cite{DBLP:journals/jmlr/ShervashidzeSLMB11}~ & ~R-convolution~       & ~No~                  &  ~Yes (Subtrees)~   & ~No~           & ~Neglect~       & ~Accommodate~\\   \hline

~CORE WL~\cite{DBLP:conf/ijcai/NikolentzosMLV18}~     & ~R-convolution~       & ~No~                  &  ~Yes (Subtrees)~   & ~No~           & ~Neglect~       & ~Accommodate~ \\   \hline

  ~SPGK~\cite{DBLP:conf/icdm/BorgwardtK05}~           & ~R-convolution~       & ~No~                  &  ~Yes (Paths)~      & ~No~           & ~Neglect~       & ~Neglect~\\  \hline

  ~CORE SP~\cite{DBLP:conf/ijcai/NikolentzosMLV18}~   & ~R-convolution~       & ~No~                  &  ~Yes (Paths)~      & ~No~           & ~Neglect~       & ~Neglect~ \\  \hline

  ~PMGK~\cite{DBLP:conf/aaai/NikolentzosMV17}~        & ~R-convolution~       & ~No~                  &  ~Yes (Vertices)~   & ~No~           & ~Neglect~       & ~Neglect~\\ \hline

  ~CORE PM~\cite{DBLP:conf/ijcai/NikolentzosMLV18}~   & ~R-convolution~       & ~No~                  &  ~Yes (Vertices)~   & ~No~           & ~Neglect~       & ~Neglect~\\ \hline

  ~RWGK~\cite{DBLP:conf/icml/KashimaTI03}~            & ~R-convolution~       & ~No~                   &  ~Yes (Walks)~     & ~No~           & ~Neglect~       & ~Accommodate~ \\ \hline

\end{tabular}
\begin{tablenotes}
\item[1] $-$: the definition of the this kernel does not refer to this property.
\end{tablenotes}
} \vspace{-0pt}
\end{table*}

\textbf{Experimental Results and Analysis:} Table~\ref{T:ClassificationGK} indicates that the proposed QESK kernel is able to outperform the alternative graph kernels on five of the nine datasets. Although, the classification accuracies of the proposed QESK kernel on the MUTAG, IMDB-B, IMDB-M and RED-B datasets are not the best, the QESK kernel is still competitive and outperforms most of these alternative kernel methods on the four datasets. The reasons of the effectiveness for the QESK kernel are twofold.

First, Table~\ref{T:GKInfor} indicates that the JTQK, GCGK, WLSK, CORE WL, SPGK, PMGK, CORE PM and RWGK kernels are all instances of R-convolution graph kernels. Thus, these kernels suffer from the same common theoretical drawbacks discussed in Section~\ref{s1.2} and Section~\ref{s3.3}, i.e., (a) cannot accommodate the effects of unshared substructures, (b) cannot distinguish structural distinctions between isomorphic substructures, and (c) restricted on local substructures of limited sizes. By contrast, the QESK kernel computes quantum-based entropic subtree representations associated with CTQW to capture complicated global and local structure information. Moreover, the similarity measure of the QESK kernel can discriminate the distinctions between isomorphic subtrees and accommodate the effects of unshared subtrees between graphs. As a result, the proposed QESK kernel significantly overcome the shortcomings of existing R-convolution kernels.

Second, although the QJSK kernel can reflect complicated structural characteristics of graphs through the CTQW. Moreover, the QJSK kernel can also capture global structural information of graphs, by measuring the Quantum Jensen-Shannon Divergence (QJSD) between the von Neumann entropy of global graphs, that is computed based on the density matrix of the CTQW. Unfortunately, unlike the proposed QESK kernel, Table~\ref{T:GKInfor} indicates that the QSJK kernel fails to capture either local structural information or vertex attributed information. Moreover, the QJSK kernel can only reflect single-viewed global structural information. By contrast, for the proposed QESK kernel, the associated the AMM matrix of the CTQW can compute different quantum-based Shannon entropies for different vertices, reflecting multi-viewed global structural information.

In summary, the proposed QESK kernel is more effective than the alternative kernels, and the strategy of computing the entropic subtree patters associated with the CTQW can significantly improve the performance of graph kernels.

\begin{table*}
\centering {
% \tiny
\scriptsize
%\footnotesize
\caption{Classification Accuracies (In $\%$ $\pm$ Standard Error) for Graph Kernels.}\label{T:ClassificationGK}
\vspace{-0pt}
\begin{tabular}{|c|c|c|c|c|c|c|c|c|c|}

  \hline
 ~Datasets~& ~MUTAG  ~        & ~NCI1~           & ~PROTEINS~         & ~D\&D~              & ~PTC(MR)~  & ~COLLAB~        & ~IMDB-B~        & ~IMDB-M~  & ~RED-B~\\ \hline \hline

  ~\textbf{QESK}~   & ~$87.02 \small{\pm0.94}$~& ~$\textbf{85.72}\small{\pm0.12}$~ & ~$\textbf{75.25}{\pm0.21}$~   &~$\textbf{80.11}\small{\pm0.29}$~ & ~$\textbf{63.26}\small{\pm0.55}$~   & ~$\textbf{79.75}\small{\pm0.16}$~  &~$73.69\small{\pm0.38}$~    &  ~$50.68\small{\pm0.27}$~ & ~$83.90\small{\pm0.11}$\\ \hline

  ~QJSK~   & ~$82.72 \small{\pm0.44}$~& ~$69.09\small{\pm0.20}$~ & ~$71.79$~   &  ~$77.68\small{\pm0.31}$~   & ~$56.70\small{\pm0.49}$~  &~$-$~       &~$62.10$~          &  ~$43.24$~ & ~$-$\\ \hline

  ~JTQK~   & ~$85.50 \small{\pm0.55}$~& ~$85.32\small{\pm0.14}$~ & ~$72.86\small{\pm0.41}$~   &  ~$79.89\small{\pm0.32}$~   & ~$58.50\small{\pm0.39}$~  &~$76.85\small{\pm0.40}$~       &~$72.45\small{\pm0.81}$~          &  ~$50.33\small{\pm0.49}$~ & ~$77.60\small{\pm0.35}$\\ \hline

  ~GCGK~   & ~$81.66\small{\pm2.11} $~ &~$62.28\small{\pm0.29}$~ & ~$71.67\small{\pm0.55}$~   &  ~$78.45\small{\pm0.26}$~   & ~$52.26\small{\pm1.41} $~ &~$72.83\small{\pm0.28}$~       &~$65.87\small{\pm0.98}$~          &  ~$45.42\small{\pm0.87}$~ & ~$77.34\small{\pm0.18}$\\ \hline

  ~WLSK~   & ~$82.88\small{\pm0.57} $~ &~$84.77\small{\pm0.13}$~ & ~$73.52\small{\pm0.43}$~   &  ~$79.78\small{\pm0.36}$~   & ~$58.26\small{\pm0.47} $~ &~$77.39\small{\pm0.35}$~       &~$71.88\small{\pm0.77}$~          &  ~$49.50\small{\pm0.49}$~ & ~$76.56\small{\pm0.30}$\\   \hline

 ~CORE WL~   & ~$87.47\small{\pm1.08} $~ &~$85.01\small{\pm0.19}$~ & ~$-$~   &  ~$79.24\small{\pm0.34}$~           & ~$59.43\small{\pm1.20} $~ &~$-$~        &~$\textbf{74.02}\small{\pm0.42}$~          &  ~$\textbf{51.35}\small{\pm0.48}$~ & ~$78.02\small{\pm0.23}$\\   \hline

  ~SPGK~   & ~$83.38\small{\pm0.81} $~ &~$74.21\small{\pm0.30}$~ & ~$75.10\small{\pm0.50}$~   &  ~$78.45\small{\pm0.26}$~   & ~$55.52\small{\pm0.46} $~&~$58.80\small{\pm0.2}$~       &~$71.26\small{\pm1.04}$~          &  ~$51.33\small{\pm0.57}$~  & ~$84.20\small{\pm0.70}$\\  \hline

  ~CORE SP~   & ~$\textbf{88.29}\small{\pm1.55} $~ &~$73.46\small{\pm0.32}$~ & ~$-$~           &  ~$77.30\small{\pm0.80}$~   & ~$59.06\small{\pm0.93} $~&~$-$~                 &~$72.62\small{\pm0.59}$~          &  ~$49.43\small{\pm0.42}$~  & ~$\textbf{90.84}\small{\pm0.14}$\\  \hline

  ~PMGK~   & ~$86.67\small{\pm0.60} $~ &~$72.91\small{\pm0.53}$~ & ~$-$~           &  ~$77.78\small{\pm0.48}$~   & ~$60.22\small{\pm0.86} $~&~$-$~                 &~$68.53\small{\pm0.61}$~          &  ~$45.75\small{\pm0.66}$~  & ~$82.70\small{\pm0.68}$\\  \hline

  ~CORE PM~   & ~$87.19\small{\pm1.47} $~ &~$74.90\small{\pm0.45}$~ & ~$-$~           &  ~$77.72\small{\pm0.71}$~   & ~$61.13\small{\pm1.44} $~&~$-$~                 &~$71.04\small{\pm0.64}$~          &  ~$48.30\small{\pm1.01}$~  & ~$87.39\small{\pm0.55}$\\  \hline

  ~RWGK~   & ~$80.77\small{\pm0.72} $~ &~$63.34\small{\pm0.27}$~       &~$74.20\small{\pm0.40}$~          &  ~$71.70\small{\pm0.47}$~         & ~$55.91\small{\pm0.37} $~  &~$-$~       &~$67.94\small{\pm0.77}$~          &  ~$46.72\small{\pm0.30}$~ & ~$-$\\ \hline

\end{tabular}
\begin{tablenotes}
\item[1] $-$: means that the method on this dataset was not reported by original authors or references.
\end{tablenotes}
} \vspace{-0pt}
\end{table*}

\subsection{Experimental Comparisons with Graph Deep Learning}

\begin{table*}
\centering {
% \tiny
 \scriptsize
%\footnotesize
\caption{Classification Accuracies (In $\%$ $\pm$ Standard Error) for Graph Deep Learning Methods.}\label{T:ClassificationGCNN}
\vspace{-0pt}

\begin{tabular}{|c|c|c|c|c|c|c|c|c|c|}

  \hline
 ~Datasets~& ~MUTAG  ~       & ~NCI1~         & ~PROTEINS~      & ~D\&D~          & ~PTC(MR)~        & ~COLLAB~        & ~IMDB-B~        & ~IMDB-M~  & ~RED-B~    \\ \hline \hline

  ~\textbf{QESK}~   & ~$87.02 \small{\pm0.94}$~& ~$\textbf{85.72}\small{\pm0.12}$~ & ~$75.25{\pm0.21}$~   &~$\textbf{80.11}\small{\pm0.29}$~ & ~$\textbf{63.26}\small{\pm0.55}$~   & ~$\textbf{79.75}\small{\pm0.16}$~  &~$\textbf{73.69}\small{\pm0.38}$~    &  ~$\textbf{50.68}\small{\pm0.27}$~ & ~$83.90\small{\pm0.11}$\\ \hline

  ~DGCNN~  & ~$85.83\small{\pm1.66}$~&~$74.44\small{\pm0.47}$~& ~$\textbf{75.54}\small{\pm0.94}$~& ~$79.37\small{\pm0.94}$~& ~$58.59\small{\pm2.47}$~ & ~$73.76\small{\pm0.49}$ ~& ~$70.03\small{\pm0.86}$ & ~$47.83\small{\pm0.85}$   & ~$76.02\small{\pm1.73}$\\ \hline

  ~PSGCNN~   & ~$\textbf{88.95}\small{\pm4.37}$~&~$76.34\small{\pm1.68}$~& ~$75.00\small{\pm2.51}$~& ~$76.27\small{\pm2.64}$~& ~$62.29\small{\pm5.68}$~ & ~$72.60\small{\pm2.15}$ ~& ~$71.00\small{\pm2.29}$ & ~$45.23\small{\pm2.84}$  & ~$\textbf{86.30}\small{\pm1.58}$\\ \hline

  ~DCNN~   & ~$66.98$~       &~$56.61\small{\pm1.04}$~& ~$61.29\small{\pm1.60}$~& ~$58.09\small{\pm0.53}$~& ~$56.60$~   & ~$52.11\small{\pm0.71}$ ~& ~$49.06\small{\pm1.37}$ & ~$33.49\small{\pm1.42}$    & ~$-$\\ \hline

  ~DGK~   & ~$82.66\small{\pm1.45}$~ &~$62.48\small{\pm0.25}$~& ~$71.68\small{\pm0.50}$~& ~$78.50\small{\pm0.22}$~    & ~$57.32\small{\pm1.13}$~ & ~$73.09\small{\pm0.25}$~& ~$66.96\small{\pm0.56}$ & ~$44.55\small{\pm0.52}$    & ~$78.30\small{\pm0.30}$\\ \hline

 % ~AWE~& ~$87.87\small{\pm9.76}$~ &~$-$~          & ~$-$~           & ~$71.51\small{\pm4.02}$~    & ~$-$~          & ~$70.99\small{\pm1.49}$~& ~$73.13\small{\pm3.28}$ & ~$51.58\small{\pm4.66}$    & ~$82.97\small{\pm2.86}$\\ \hline

  ~ECCN~    & ~$76.11$~       &~$76.82$~       & ~$72.65$~           & ~$74.10$~       & ~$-$~            & ~$67.79$~            & ~$-$            & ~$-$    & ~$-$\\ \hline

\end{tabular}
\begin{tablenotes}
\item[1] $-$: means that the method on this dataset was not reported by original authors or references.
\end{tablenotes}
} \vspace{-0pt}
\end{table*}

\noindent\textbf{Experimental Setups:} We also compare the performance of the proposed QESK kernel with some graph deep learning methods, including (1) the Deep Graph Convolutional Neural Network (DGCNN)~\cite{DBLP:conf/aaai/ZhangCNC18}, (2) the PATCHY-SAN based Convolutional Neural Network for graphs (PSGCNN)~\cite{DBLP:conf/icml/NiepertAK16}, (3) the Diffusion Convolutional Neural Network (DCNN)~\cite{DBLP:conf/nips/AtwoodT16}, (4) the Deep Graph Kernel (DGK)~\cite{DBLP:conf/kdd/YanardagV15}, and (5) the Edge-conditional Convolutional Networks (ECCN)~\cite{DBLP:conf/cvpr/SimonovskyK17}. Because these alternative graph deep learning methods follow the same experimental setup with the proposed QESK kernel, we directly report the classification accuracies from the original references in Table~\ref{T:ClassificationGCNN}. Note that, both the PSGCNN and ECCN models can accommodate edge attributes. However, since most datasets and the alternative methods do not leverage edge attributes, we only report the results of the two models associated with vertex attributes.

\textbf{Experimental Results and Analysis:} Table~\ref{T:ClassificationGCNN} indicates that the proposed QESK kernel is able to outperform the alternative graph deep learning approaches on six of the nine datasets. Although, the classification accuracies of the proposed QESK kernel on the MUTAG, PROTEINS and RED-B datasets are not the best, the QESK kernel is still competitive and outperforms most of these alternative methods on these datasets. In fact, the C-SVM associated with the graph kernel is essentially a kind of learning method based on the shallow learning strategy, and usually has lower performance than deep learning. However, the proposed QESK can still outperform all the alternative graph deep learning methods, demonstrating the effectiveness.

One reason for the effectiveness is that the alternative DGCNN, PSGCNN and DCN models have been shown their theoretical relationships to the classical WL Tree-Index method~\cite{DBLP:conf/aaai/ZhangCNC18}. As typical instances of Graph Neural Networks (GNN), these models all rely on propagating vertex attributed information to neighborhood vertices, and this computational procedure is similar to the WL Tree-Index method that takes the union operation for the vertex attributes of adjacent vertices. Thus, the structural representation of these GNN models may be limited by the WL Tree-Index method. By contrast, the quantum-based entropic subtree representations with the WL Tree-Index method can significantly strengthen the structural representation through the AMM matrix of the CTQW, that can capture complicated structural information of graphs. As a result, the QESK kernel can better reflect the structural information of graphs than these GNN models related to the original WL Tree-Index method. This evaluation indicates that the strategy of computing the entropic subtree patters associated with the CTQW can significantly improve the performance of graph kernels.

\section{Conclusion and Future Work}\label{s5}

In this work, we have developed a novel QESK kernel for graph classification. The proposed kernel is defined by measuring the similarity between graphs associated with their entropic subtree representations, that are computed based on the AMM matrix of the CTQW as well as the WL Tree-Index method. Unlike state-of-the-art R-convolution graph kernels, the proposed QESK kernel can not only reflect the effect of unshared substructures between graphs, but also identify the structural distinctions between isomorphic substructures in terms of the structural arrangement of global graph structures. Furthermore, the proposed QESK kernel can simultaneously reflect global and local structural information of graphs. As a result, the proposed QESK kernel can theoretically address the shortcoming of existing R-convolution kernels, explaining the effectiveness. The experimental evaluations indicate the effectiveness.

In the previous work~\cite{DBLP:conf/icpr/BaiRH14}, we have proposed a hypergraph kernel associated with the WL Tree-Index method on directed line graphs~\cite{DBLP:journals/pr/BaiEH16}. This kernel not only accommodates hypergraphs but also restricts the tottering problems arising in the WL Tree-Index method through the directed edge. The future work is to combine the idea of the proposed QESK kernel and the hypergraph kernel together, developing novel quantum-based subtree kernels for hypergraphs.

\section*{Acknowledgments}
This work is supported by the National Natural Science Foundation of China under Grants T2122020, 61976235, and 61602535. Corresponding Author: Linxin Cui (cuilixin@cufe.edu.cn).

% was supervised by Dr. Lu Bai and Dr. Lixin Cui for his M.sc degree and

%-------------------------------------------------------------------------

\balance

%-------------------------------------------------------------------------
%\nocite{ex1,ex2}
%\bibliographystyle{latex12}
%\bibliography{XBib}

\bibliographystyle{IEEEtran}
\bibliography{example_paper}

\end{document}